\documentclass[10pt,twocolumn,letterpaper]{article}

\usepackage[pagenumbers]{cvpr} 

\usepackage{graphicx}
\usepackage{amsmath}
\usepackage{amssymb}
\usepackage{booktabs}
\usepackage{multirow}
\usepackage{indentfirst}
\usepackage{times}
\usepackage{epsfig}
\usepackage{subcaption}

\usepackage[table,xcdraw]{xcolor}

\usepackage[pagebackref,breaklinks,colorlinks]{hyperref}

\usepackage[capitalize]{cleveref}
\crefname{section}{Sec.}{Secs.}
\Crefname{section}{Section}{Sections}
\Crefname{table}{Table}{Tables}
\crefname{table}{Tab.}{Tabs.}

\def\cvprPaperID{6079} 
\def\confName{CVPR}
\def\confYear{2022}

\begin{document}

\title{Rethinking Re-Sampling in Imbalanced Semi-Supervised Learning}

\author{
	Ju He$^{1}$
	\;\; Adam Kortylewski$^1$
	\;\; Shaokang Yang$^2$\\
	\;\; Shuai Liu$^2$
	\;\; Cheng Yang$^2$
	\;\; Changhu Wang$^2$
	\;\; Alan Yuille$^1$\\
	$^1$Johns Hopkins University \;\;  $^2$ByteDance Inc.\\	
}

\maketitle

\begin{abstract}
Semi-Supervised Learning (SSL) has shown its strong ability in utilizing unlabeled data when labeled data is scarce. However, most SSL algorithms work under the assumption that the class distributions are balanced in both training and test sets. In this work, we consider the problem of SSL on class-imbalanced data, which better reflects real-world situations. In particular, we decouple the training of the representation and the classifier, and systematically investigate the effects of different data re-sampling techniques when training the whole network including a classifier as well as  fine-tuning the feature extractor only. We find that data re-sampling is of critical importance to learn a good classifier as it increases the accuracy of the pseudo-labels, in particular for the minority classes in the unlabeled data. Interestingly, we find that accurate pseudo-labels do not help when training the feature extractor, rather contrariwise, data re-sampling harms the training of the feature extractor. This finding is against the general intuition that wrong pseudo-labels always harm the model performance in SSL. Based on these findings, we suggest to re-think the current paradigm of having a single data re-sampling strategy and develop a simple yet highly effective Bi-Sampling (BiS) strategy for SSL on class-imbalanced data. BiS implements two different re-sampling strategies for training the feature extractor and the classifier and integrates this decoupled training into an end-to-end framework. In particular, BiS progressively changes the data distribution during training such that in the beginning the feature extractor is trained effectively, while towards the end of the training the data is re-balanced such that the classifier is trained reliably. We benchmark our proposed bi-sampling strategy extensively on popular datasets and achieve state-of-the-art performances. Code is available at \url{https://github.com/TACJu/Bi-Sampling}.
\end{abstract}
\section{Introduction}

Deep neural networks \cite{he2015deep, ren2016faster, long2015fully} have revolutionised numerous fields in recent years and achieved huge success due to the existence of many high-quality, large-scale labeled datasets \cite{russakovsky2015imagenet, lin2015microsoft}. Yet, collecting large amounts of labeled data is expensive and sometimes even impossible in some scenarios due to the labor-intensive annotation along with experts knowledge requirement. For example, labeling large-scale, long-range video sequential \cite{abuelhaija2016youtube8m} for video classification task and medical datasets \cite{johnson2008accuracy, zbontar2018fastMRI} which requires experts involved to develop algorithms for detecting tumors might be two examples to well illustrate the high cost needed for annotating these datasets. In contrast, unlabeled data are usually much easier to acquire and can also be exploited to improve model performance.

\begin{figure*}[]
    \centering
    \includegraphics[width=\linewidth,height=4.5cm]{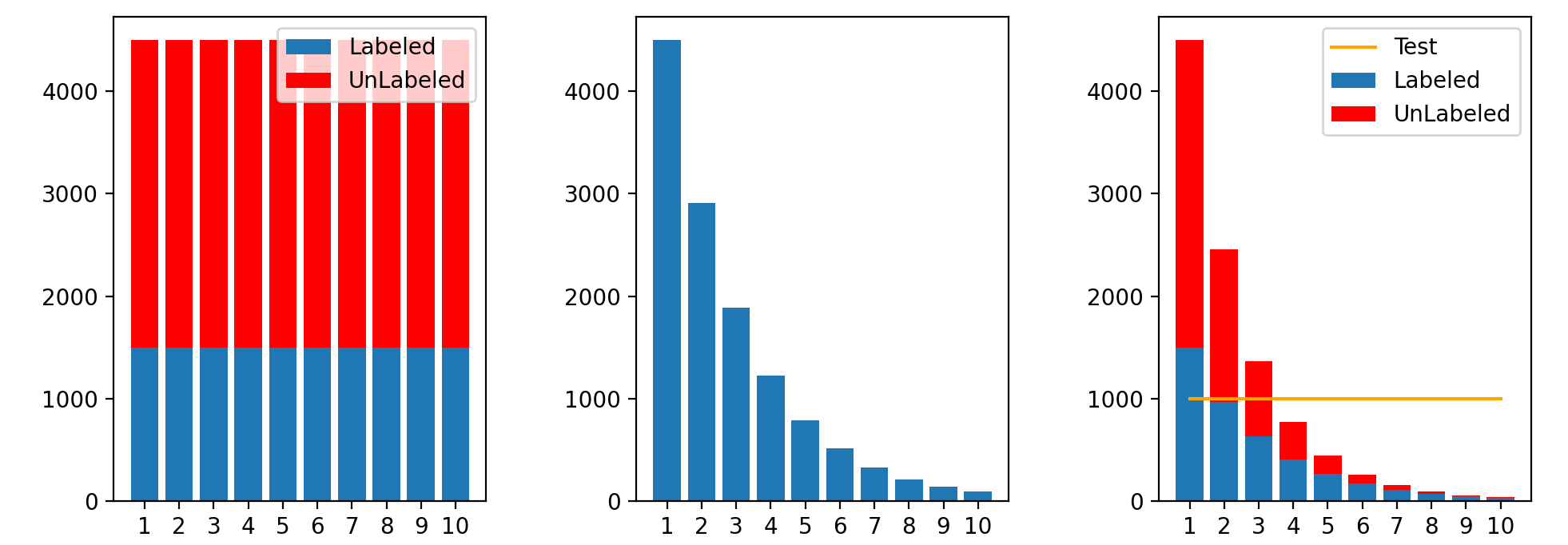}
    \caption{\textbf{Left}: Data distribution of general semi-supervised learning. \textbf{Middle}: Data distribution of imbalanced classification. \textbf{Right:} Data distribution of imbalanced semi-supervised learning where both labeled and unlabeled sets are class-imbalanced but test sets are balanced.}
    \label{fig:intro}
\end{figure*}

A powerful approach for training models on large amounts of data with only part of it being labeled is semi-supervised learning (SSL). However, most existing works in SSL \cite{berthelot2019mixmatch, berthelot2019remixmatch, sohn2020fixmatch} work under the assumption that both labeled and unlabeled data have a balanced class distribution, i.e., each class has roughly the same number of samples (Figure \ref{fig:intro} Left). SSL methods are effective under this assumption but have not been thoroughly evaluated when this assumption does not hold.

In contrast, compared to well-curated vision datasets which exhibit uniform distributions of class labels, real-world data usually has skewed distributions with a long tail, where a few classes (a.k.a. head classes) have most of the data points, while many classes (a.k.a. tail classes) only have rarely few samples (Figure \ref{fig:intro} Middle). It has been widely observed \cite{NIPS2017_147ebe63, zhong2019unequal} that models directly trained on such long-tailed distributions are biased towards head classes and away from tail classes thus we need to balance the bias. 
A simple yet efficient idea for tackling long-tailed distributions is re-sampling, i.e. sampling more data from the tailed classes to account for their scarcity. Effects of different degrees of re-sampling strength have been studied in previous works \cite{mahajan2018exploring, shen2016relay} and thus result in different re-sampling strategies. Recent works \cite{kang2019decoupling, zhou2020bbn} point out that in the supervised setting, it is better to first train the feature extractor and classifier jointly without any re-sampling strategy followed by only fine-tuning the classifier with re-sampling strategies. This process is called decoupling since it suggests that learning representation and classifier requires different balancing strategies.

While the aforementioned methods have greatly improved the model performance when facing imbalanced data in a fully supervised setting, semi-supervised learning on imbalanced data has not been a lot of attention yet. 
In fact, in a semi-supervised learning setting, balancing the skewed data distribution becomes even more challenging because most of the data is not labelled and the pseudo labels generated by SSL algorithms are unreliable. 
A few recent works \cite{hyun2020class, kim2020distribution, wei2021crest} tackle this problem in a joint learning manner and find out that these pseudo labels are biased towards head classes and subsequent training with such problematic pseudo labels may even intensify the bias which further degrades the model performance. However, they do not decouple the learning process and thus fail to analyze the effect of balancing strategies for the classifier and the feature extractor individually. 
As a result, all prior works only use a single fixed balancing strategy during the whole training process which is shown to be sub-optimal in our experiments.

In this paper, we investigate SSL in the context of long-tailed data distribution where both labeled and unlabeled data have the same imbalanced class ratio (Figure \ref{fig:intro} Right).
We conduct the first systematic study of different re-sampling strategies for SSL when training feature extractor and classifier jointly, as well as in a decoupled learning scheme, where we fine-tune the classifier of jointly trained models with a different re-sampling strategy.
Interestingly, we find out that though producing much more biased pseudo labels for unlabeled data, models trained without re-sampling learn the best feature representation and get top performance after fine-tuning a class-balanced classifier.

This phenomenon indicates that re-sampling actually harms the learning of representations thus suggesting that different sampling strategies are required for training the representation and classifier. Based on that finding, we develop a novel yet simple two-sampler method named \textit{BiS} which helps the model to first learn a good representation and then gradually transfer its focus to obtain a more balanced classifier that can be trained in an end-to-end manner. To be specific, two data samplers with different re-sampling strategies construct the training batches of the model together where the sample ratio of the two data samplers progressively changes during the training process. Extensive experiments on CIFAR-10 and CIFAR-100 datasets are conducted to evaluate our method under different experimental settings (e.g. various unlabeled data ratio and imbalanced data ratio) where our method shows consistent improvements compared to previous state-of-the-art methods.

In summary, we make following contributions in this paper:

\begin{enumerate}
    \item We systematically analyze the effect of different re-sampling strategies for SSL on class-imbalanced data in a joint and a decoupled network training scenarios.
    \item We find that the biased pseudo labels produced by SSL algorithms do harm the classifier, but do not harm the capabilities of the feature extractor, on the contrary, re-sampling reduces the quality of the feature representations. 
    \item Our experimental findings enable us to develop a novel training strategy called \textit{Bi-Sampling} (BiS), which integrates two different re-sampling strategies for training the feature extractor and the classifier into a uniform end-to-end framework. Models trained with BiS achieve state-of-the-art performances on all imbalanced semi-supervised benchmarks.
\end{enumerate}
\section{Related Works}
\label{sec:related}
\subsection{Semi-supervised learning}
Semi-supervised learning aims at learning from both labeled data and unlabeled data. Many of existing methods \cite{berthelot2019mixmatch, berthelot2019remixmatch, sohn2020fixmatch} use pseudo labeling and consistency regularization. In particular, pseudo labeling \cite{mclachlan1975iterative, lee2013pseudo} leverages the idea of using the model itself to produce 'hard' artificial labels for unlabeled data. A manual threshold is adopted to decide whether to retain the artificial labels or not. This idea is closely related to entropy minimization \cite{grandvalet2005semi, sajjadi2016mutual} which encourages the model predictions to be low-entropy (i.e. high confidence) on unlabeled data. Consistency regularization \cite{bachman2014learning, sajjadi2016regularization} relies on the assumption that the model should output similar predictions when fed perturbed versions of the same image. The state-of-the-art SSL algorithm FixMatch \cite{sohn2020fixmatch} integrates these basic techniques all and achieves excellent results. However, all of the aforementioned works assume that the labeled data and unlabeled data share the same uniform distribution. In a class-imbalanced setting \cite{kim2020distribution, wei2021crest} the pseudo labels are significantly biased and therefore in this work we analyze the effect of this phenomenon.

\subsection{Class-imbalanced supervised learning}
Class-imbalanced supervised learning has drawn increasing attention due to its relevance in real-world applications. Recent studies have mainly pursued three directions: re-sampling \cite{zhou2020bbn, kang2019decoupling}, re-weighting \cite{cui2019class, cao2019learning} and transfer learning \cite{kim2020m2m, Liu_2020_CVPR}. Re-sampling methods manually sample the data according to a pre-defined distribution to get a more balanced training set and re-weighting methods assign higher weights to tail classes instances to balance the overall contribution while transfer learning aims at transferring knowledge from head classes to tail classes. Recent work \cite{kang2019decoupling} shows that in a decoupled learning scenario, a simple re-sampling strategy can achieve state-of-the-art performance compared to more complicated counterparts. However, these methods highly rely on the labels of the data and their performance have not been tested under SSL scenarios extensively yet.

\subsection{Class-imbalanced semi-supervised learning}
Some very recent works started focusing on class-imbalanced semi-supervised learning setting, as it more accurately describes the real-world data distribution. Yang and Xu \cite{yang2020rethinking} claimed that leveraging unlabeled data either by semi-supervised learning or self-supervised learning can both help to alleviate the bias problem in class-imbalanced learning. Hyun et al. \cite{hyun2020class} proposed a suppressed consistency loss on minority classes to boost the performance. Kim et al. \cite{kim2020distribution} proposed Distribution Aligning Refinery (DARP) to conduct a quick alignment between the predictions and desired distribution through solving convex optimization. Wei et al. \cite{wei2021crest} found that the raw SSL methods usually have high recall and low precision for head classes while the reverse is true for the tail classes and further proposed a reverse sampling method for unlabeled data based on that. In this work, we further analyze this interesting phenomenon and propose a simple yet effective Bi-Sampling strategy to boost the performance based on that.
\section{An Empirical Study of Re-Sampling and Decoupling in Imbalanced SSL}
\label{sec:empirical}
In this section, we first define the problem setup of class-imbalanced semi-supervised learning and introduce different data sampling strategies. Then we present a detailed analysis on different strategies of sampling data when training the models in a joint manner followed by a discussion on the effects of re-sampling at learning representations.

\begin{table*}[ht]
    \centering
    \setlength{\abovecaptionskip}{0pt}   
    \setlength{\belowcaptionskip}{10pt}
    \caption{Empirical study of different sampling strategies for the labeled and unlabeled data in class-imbalanced SSL using a joint learning scheme. We report the classification accuracy (\%) on CIFAR-10 with varying class-imbalance ratio $\lambda$ and varying ratio of unlabeled data $\beta$. The best results are marked in bold.}
    \begin{tabular}{|c|c|ccc|ccc|}
        \hline
        \multirow{2}{*}{\begin{tabular}[c]{@{}c@{}}Labeled Data\\ Sampling\end{tabular}} & \multirow{2}{*}{\begin{tabular}[c]{@{}c@{}}Unlabeled Data\\ Sampling\end{tabular}} & \multicolumn{6}{c|}{Imbalanced Ratio $\lambda$ / Unlabeled Ratio $\beta$} \\ \cline{3-8} 
         &  & 50/1 & 100/1 & 150/1 & 50/2 & 100/2 & 150/2 \\ \hline
        Random & Random & 75.8 & 67.3 & 62.4 & 80.8 & 73.9 & 68.3 \\
        Random & Mean & 78.4 & 71.7 & 66.2 & 81.3 & 74.9 & 70.7 \\
        Random & Reverse & 78.6 & 70.5 & 64.6 & 78.5 & 71.6 & 67.0 \\
        Mean & Random & 77.8 & 70.0 & 65.5 & \textbf{83.1} & 76.2 & 71.4 \\
        Mean & Mean & \textbf{79.5} & 71.8 & 68.3 & 82.5 & \textbf{76.8} & 71.5 \\
        Mean & Reverse & 79.0 & \textbf{72.3} & 68.8 & 79.8 & 73.5 & 69.4 \\
        Reverse & Random & 76.5 & 70.5 & 61.3 & 82.9 & 76.4 & 71.0 \\
        Reverse & Mean & 78.3 & 70.3 & 66.5 & 81.5 & 75.8 & \textbf{72.1} \\
        Reverse & Reverse & 78.0 & 72.0 & \textbf{68.9} & 79.9 & 73.3 & 70.3 \\ \hline
    \end{tabular}
    \label{tab:cifar10joint}
\end{table*}

\begin{figure*}[ht]
    \centering
    \includegraphics[width=\linewidth,height=6cm]{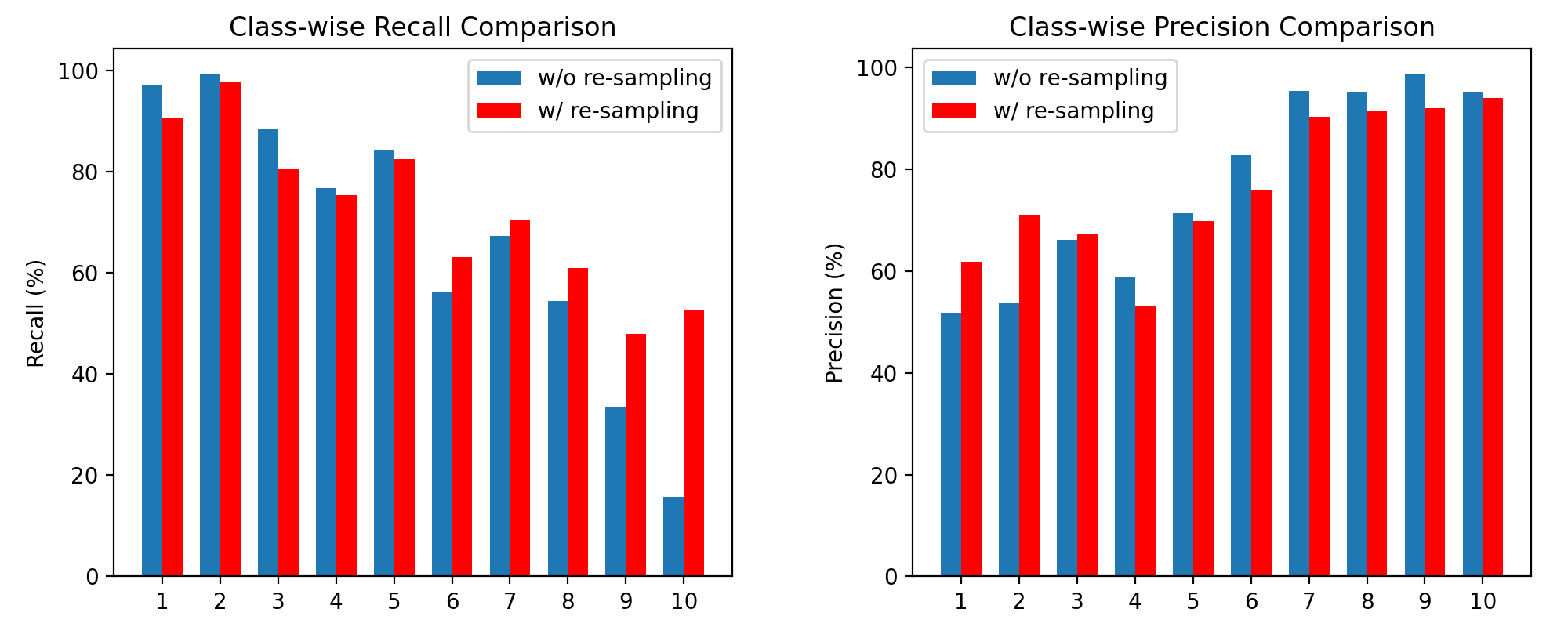}
    \caption{Bias comparison between FixMatch models w \& w/o re-sampling on CIFAR-10. \textbf{Left}: Per-class recall comparison. \textbf{Right}: Per-class precision comparison. The class index is sorted by the number of examples in descending order. The models show a descending trend in recall while an ascending trend in precision.}
    \label{fig:bias}
\end{figure*}

\subsection{Preliminaries}
\label{sec:setup}

\textbf{Problem setup.} For a K-class classification problem, there is a labeled training set $\mathbb{X}=\{(x_n, y_n) \colon n \in (1,..., N)\}$ and a unlabeled training set $\mathbb{U}=\{(u_m) \colon m \in (1,..., M)\}$, where $x_n, u_m \in \mathbb{R}^d$ are training examples and $y_n \in \{1,...,K\}$ are corresponding class labels. The number of training samples in $\mathbb{X}$ of class $k$ is denoted as $N_k$, i.e., $\sum_{k=1}^{K}N_k=N$. We assume that $\mathbb{X}$ and $\mathbb{U}$ share the same skewed distribution which is the most common situation in the real world. Without loss of generality, the classes are sorted by $N_k$ in a descending order ($N_1 \geq N_2 \geq ... \geq N_K$) and thereby we can represent the imbalanced ratio $\lambda = \frac{N_1}{N_K}$. We denote the ratio of unlabeled data to labeled data as $\beta = \frac{M}{N}$. Our model can be represented as a combination of a feature extractor $(f(x;\theta))$ and a classifier $g$. Given the class-imbalanced datasets $\mathbb{X}$ and $\mathbb{U}$, we aim at training our model $ g(f(x;\theta)) \rightarrow \{1,...,K\}$ in a semi-supervised manner to generalize well on a class-balanced test set.

\noindent \textbf{Sampling Strategies.} We focus on different sampling strategies to re-balance the data distribution for the learning process of our model. We will change the sampling strategy for both labeled dataset $X$ and the unlabeled dataset $U$. The following three sampling strategies are introduced.

\emph{Random Sampling.} The most common way and also the standard way of sampling data is to completely sample data points from the dataset randomly which we name random sampling here. Formally, under our setting, the probability that a sample from class $j$ is chosen from the dataset is $\mu_j = \frac{N_j}{\sum_{i=1}^{K}N_i}$. The long-tailed distribution of the dataset is unchanged in this situation.

\emph{Mean Sampling.} Mean sampling that assigns an equal probability $\mu_j = \frac{1}{K}$ for each class to be chosen has been shown to be better than random sampling when facing skewed data distribution \cite{shen2016relay, mahajan2018exploring} since the tail classes will have more chances to be sampled.

\emph{Reverse Sampling.} To face the challenges that sometimes mean sampling is not enough to solve the bias introduced with the skewed distribution. We further propose a more aggressive sampling strategy named reverse sampling that gives a higher chance for tail classes to be chosen. The probability of sampling a data point of class $j$ is given by $\mu_j = \frac{\frac{1}{N_j}}{\sum_{i=1}^{K}\frac{1}{N_i}}$. In this way, the data distribution is reversed.

\noindent \textbf{Utilizing Unlabeled Data.} Instead of utilizing all the samples in the unlabeled dataset $\mathbb{U}$, we propose to only use a subset $\mathbb{S}$ of $\mathbb{U}$ to expand labeled dataset $\mathbb{X}$. In details, the choice of subset $\mathbb{S}$ contains two steps. First we rule out all the data that does not meet the quality threshold of SSL algorithms (e.g. FixMatch has a confidence threshold on the classification results of the augmented samples). Then with the chosen re-sampling strategy, we can calculate the probability $\mu_j$ for an unlabeled data sample of pseudo label $j$ with the formula above. We further add a hyper-parameter $q$ ($q \ge 0$) to control the strength of re-sampling thus the final keep probability adding the sample to $\mathbb{S}$ can be expressed as ${\mu_j}^q$ where $q$ can effectively tune the keep probability of each class. For example, when $q = 0$, then the probability for all classes becomes $1$ thus the quality threshold becomes the only factor deciding $\mathbb{S}$. While when $q = 1$, the keep probability is just the same as that for labeled data. Unlike \cite{wei2021crest}, we do not manually select the most confident pseudo-labeled samples of each class because the biased pseudo label confidence is usually unreliable and the manual selection will require an offline process which complicates the whole pipeline.

\noindent \textbf{Joint and Decoupled Training Schemes.} Assume that our model is constructed with a feature extractor $f$ along with a classifier $g$. Joint training means that we use a standard cross-entropy loss to update $f$ and $g$ together. While following \cite{kang2019decoupling}, decoupled training refers to the procedure that after obtaining a jointly trained model, we freeze the parameters of $f$ and only fine-tune the classifier $g$.

\begin{table*}[ht]
    \centering
    \small
    \setlength{\abovecaptionskip}{0pt}   
    \setlength{\belowcaptionskip}{10pt}
    \caption{Classification accuracy (\%) after fine-tuning the models from Table \ref{tab:cifar10joint}. We freeze the feature extractor and train the classifier with a mean sampling strategy on CIFAR-10. Performance variation is shown in brackets. Best results are bolded. Results with the greatest improvement are underlined.}
    \begin{tabular}{|c|ccc|ccc|}
        \hline
        \multirow{2}{*}{\begin{tabular}[c]{@{}c@{}}Sampling Strategy for\\ labeled / unlabeled data\end{tabular}} & \multicolumn{6}{c|}{Imbalanced Ratio $\lambda$ / Unlabeled Ratio $\beta$} \\ \cline{2-7} 
         & 50/1 & 100/1 & 150/1 & 50/2 & 100/2 & 150/2 \\ \hline
        Random / Random & {\underline{\textbf{80.3(+4.5)}}} & {\underline{\textbf{74.6(+7.3)}}} & {\underline{69.0(+6.6)}} & {\underline{\textbf{84.6(+3.8)}}} & {\underline{\textbf{78.5(+4.6)}}} & {\underline{74.9(+6.6)}} \\
        Random / Mean & 80.2(+1.8) & 74.1(+2.4) & \textbf{69.5(+3.3)} & 83.2(+1.9) & 77.9(+3.0) & \textbf{75.2(+4.5)} \\
        Random / Reverse & 79.3(+0.7) & 72.7(+2.2) & 67.1(+2.5) & 80.5(+2.0) & 75.5(+3.9) & 73.4(+6.4) \\
        Mean / Random & 78.9(+1.1) & 73.0(+3.0) & 66.6(+1.1) & 84.1(+1.0) & 77.4(+1.2) & 72.4(+1.0) \\
        Mean / Mean & 79.1(-0.4) & 73.6(+1.8) & 68.9(+0.6) & 82.8(+0.3) & 77.0(+0.2) & 72.7(+1.2) \\
        Mean / Reverse & 79.1(+0.1) & 73.5(+1.2) & 69.0(+0.2) & 79.7(-0.1) & 73.6(+0.1) & 69.7(+0.3) \\
        Reverse / Random & 76.0(-0.5) & 71.5(+1.0) & 61.6(+0.3) & 82.7(-0.2) & 76.4(+0.0) & 70.7(-0.3) \\
        Reverse / Mean & 77.3(-0.7) & 68.9(-1.4) & 64.9(-1.6) & 80.5(-1.0) & 72.6(-3.2) & 70.2(-1.9) \\
        Reverse / Reverse & 77.6(-0.4) & 70.5(-1.5) & 65.3(-3.6) & 78.3(-1.6) & 70.8(-2.5) & 66.3(-4.0) \\ \hline
    \end{tabular}
    \label{tab:cifar10decoupled}
\end{table*}

\subsection{Data re-sampling in a joint learning process}
\label{sec:joint}
\textbf{Experimental Setup.} We conduct thorough experiments on semi-supervised long-tailed datasets with different imbalanced ratio $\lambda$ and unlabeled ratio $\beta$. To be specific, we follow the procedure of DARP \cite{kim2020distribution} to build the desired version of CIFAR-10 \cite{krizhevsky2009learning} for a fair comparison with prior works and further extend it to CIFAR-100. We present results on CIFAR-10 in the main body while leaving the part of CIFAR-100 and the details of constructing such datasets in the Appendix. In all experiments, we use FixMatch \cite{sohn2020fixmatch} as our baseline model, which is the state-of-the-art method in SSL. Following \cite{kim2020distribution, wei2021crest}, we choose WRN-28 \cite{zagoruyko2016wide} as our backbone.

\noindent \textbf{Main Results.} Table \ref{tab:cifar10joint} summarizes the results on CIFAR-10 dataset with $\lambda = 50, 100, 150$ and $\beta = 1, 2$ with a joint training scheme. FixMatch with random sampling performs reasonably well on the imbalanced ratio $\lambda = 50$, but the accuracy decreases significantly with increasing $\lambda$. In contrast, proper balancing strategies can strongly boost the overall performance compared to random sampling. On average with changing re-sampling strategies FixMatch can obtain $4\%$ improvement. The absolute accuracy gain is as much as $6.5\%$ when $\lambda=150$ and $\beta=1$ with the reverse sampler for both labeled data and unlabeled data. 

\emph{Per-class Precision \& Recall Analysis.} 
To take a closer look at why the performance of the model is much lower than their balanced counterparts and to account for the reason why simple re-balancing strategies work, we analyze the per-class precision and recall in Figure \ref{fig:bias} to better understand the observed performances in Table \ref{tab:cifar10joint}. We observe that our baseline FixMatch without re-sampling achieves high recall on head classes but very low recall on tail classes. For example, the recall on the most majority class of CIFAR-10 is $97.3\%$. On the contrary, that of the most minority class is only $15.7\%$. Meanwhile, the precision of head classes and tail classes follows exactly the opposite pattern, i.e. the precision of the minority class is $95.2\%$, while that of the most majority class is only $51.9\%$. These two trends clearly show that the model without re-sampling will wrongly classify most of the tail class samples into head classes with low confidence but have high confidence on the correctly classified tail classes samples. This provides us with empirical evidence that supports the manual re-sampling to increase the frequency of tail class data during training. As a comparison, FixMatch with a mean sampler for both labeled and unlabeled data, though still subject to these general trends, greatly alleviates the biases (e.g. the most minority recall: $15.7\%\rightarrow52.8\%$, the most majority precision: $51.9\%\rightarrow61.8\%$). This explains the overall accuracy gain on the class-balanced test sets (i.e. $67.3\%\rightarrow72.3\%$).

\emph{Summary.} From our study of different re-sampling strategies in a joint learning scheme, we observe that \textbf{re-sampling helps to alleviate the model bias towards head classes and greatly improves the performance on tail classes}. While this finding aligns with the conclusion in the supervised setting, we confirm its correctness in imbalanced SSL and further analyze the potential reason.

\subsection{Data re-sampling in a decoupled learning process}
\label{sec:decouple}
In this section, we study the decoupled learning of the representation and the classifier in the context of class-imbalanced SSL. Though decoupling is shown to be effective in the supervised setting, we wonder whether it is still true when unlabeled data is involved. 
We use the models trained with different re-sampling strategies in Section \ref{sec:joint} as our pretrained models. Then, we fix the parameters of the feature extractor and randomly initialize the parameters of the classifier followed by fine-tuning it with a different balancing strategy to handle the class-balanced test sets. 

\begin{table*}[ht]
    \centering
    \setlength{\abovecaptionskip}{0pt}   
    \setlength{\belowcaptionskip}{10pt}
    \caption{Comparison of classification accuracy (\%) with previous state-of-the-art methods on class-imbalanced semi-supervised CIFAR-10 based on different SSL algorithms. The numbers are averaged over 5 different folds. Best results are bolded.}
    \begin{tabular}{|c|ccc|ccc|}
        \hline
        \multirow{2}{*}{SSL Algorithm} & \multicolumn{6}{c|}{Imbalanced Ratio $\lambda$ / Unlabeled Ratio $\beta$} \\ \cline{2-7} 
         & 50/1 & 100/1 & 150/1 & 50/2 & 100/2 & 150/2 \\ \hline
        MixMatch & $71.1_{\pm0.33}$ & $63.6_{\pm0.41}$ & $58.5_{\pm0.68}$ & $73.2_{\pm0.29}$ & $64.8_{\pm0.31}$ & $62.5_{\pm0.80}$ \\
        MixMatch + DARP + cRT & $72.1_{\pm0.61}$ & $65.6_{\pm0.47}$ & $61.6_{\pm0.69}$ & $76.0_{\pm0.18}$ & $68.6_{\pm0.83}$ & $67.0_{\pm0.77}$ \\
        MixMatch + CReST+ & $76.3_{\pm0.24}$ & $66.2_{\pm0.15}$ & $62.9_{\pm0.68}$ & $79.0_{\pm0.40}$ & $71.9_{\pm0.35}$ & $68.3_{\pm0.74}$ \\
        MixMatch + BiS & $\textbf{77.0}_{\pm0.24}$ & $\textbf{67.8}_{\pm0.47}$ & $\textbf{63.6}_{\pm0.58}$ & $\textbf{80.0}_{\pm0.29}$ & $\textbf{72.6}_{\pm0.51}$ & $\textbf{68.7}_{\pm0.48}$ \\ \hline
        FixMatch & $75.8_{\pm0.45}$ & $67.3_{\pm0.36}$ & $62.4_{\pm0.88}$ & $80.8_{\pm0.11}$ & $73.9_{\pm0.65}$ & $68.3_{\pm0.62}$ \\
        FixMatch + DARP + cRT & $78.2_{\pm0.25}$ & $68.9_{\pm0.32}$ & $66.2_{\pm0.63}$ & $82.4_{\pm0.41}$ & $76.1_{\pm0.73}$ & $71.5_{\pm0.70}$ \\
        FixMatch + CReST+ & $79.9_{\pm0.08}$ & $74.4_{\pm0.21}$ & $69.8_{\pm0.54}$ & $83.9_{\pm0.40}$ & $77.4_{\pm0.26}$ & $72.8_{\pm0.55}$ \\
        FixMatch + BiS & $\textbf{80.8}_{\pm0.07}$ & $\textbf{75.5}_{\pm0.24}$ & $\textbf{71.0}_{\pm0.72}$ & $\textbf{85.5}_{\pm0.10}$ & $\textbf{81.0}_{\pm0.36}$ & $\textbf{76.9}_{\pm0.61}$ \\ \hline
    \end{tabular}
    \label{tab:cifar10all}
\end{table*}

\begin{table*}[ht]
    \centering
    \setlength{\abovecaptionskip}{0pt}   
    \setlength{\belowcaptionskip}{10pt}
    \caption{Comparison of classification accuracy (\%) with previous state-of-the-art methods on class-imbalanced semi-supervised CIFAR-100 based on different SSL algorithms. The numbers are averaged over 5 different folds. Best results are bolded.}
    \begin{tabular}{|c|ccc|ccc|}
        \hline
        \multirow{2}{*}{SSL Algorithm} & \multicolumn{6}{c|}{Imbalanced Ratio $\lambda$ / Unlabeled Ratio $\beta$} \\ \cline{2-7} 
         & 10/1 & 20/1 & 50/1 & 10/2 & 20/2 & 50/2 \\ \hline
        MixMatch & $51.6_{\pm0.40}$ & $45.6_{\pm0.31}$ & $37.2_{\pm0.69}$ & $52.8_{\pm0.32}$ & $45.8_{\pm0.45}$ & $37.6_{\pm0.98}$ \\
        MixMatch +DARP + cRT & $53.1_{\pm0.82}$ & $47.9_{\pm1.31}$ & $41.0_{\pm1.02}$ & $55.7_{\pm0.68}$ & $51.3_{\pm0.80}$ & $44.2_{\pm1.24}$ \\
        MixMatch + CReST+ & $54.0_{\pm0.22}$ & $48.7_{\pm0.31}$ & $40.6_{\pm0.50}$ & $55.9_{\pm0.11}$ & $51.3_{\pm0.59}$ & $44.7_{\pm0.47}$ \\
        MixMatch + BiS & $\textbf{55.2}_{\pm0.14}$ & $\textbf{49.3}_{\pm0.44}$ & $\textbf{41.5}_{\pm0.35}$ & $\textbf{56.6}_{\pm0.17}$ & $\textbf{51.9}_{\pm0.32}$ & $\textbf{45.6}_{\pm0.73}$ \\ \hline
        FixMatch & $50.2_{\pm0.13}$ & $44.1_{\pm0.29}$ & $35.6_{\pm0.59}$ & $54.9_{\pm0.18}$ & $48.0_{\pm0.15}$ & $40.5_{\pm0.44}$ \\
        FixMatch + DARP + cRT & $51.8_{\pm0.86}$ & $45.9_{\pm1.12}$ & $37.8_{\pm1.47}$ & $56.2_{\pm0.60}$ & $49.3_{\pm0.72}$ & $42.3_{\pm0.93}$ \\
        FixMatch + CReST+ & $52.5_{\pm0.31}$ & $46.6_{\pm0.71}$ & $39.5_{\pm0.59}$ & $56.1_{\pm0.46}$ & $49.5_{\pm0.65}$ & $43.2_{\pm0.83}$ \\
        FixMatch + BiS & $\textbf{53.4}_{\pm0.13}$ & $\textbf{47.3}_{\pm0.42}$ & $\textbf{40.6}_{\pm0.65}$ & $\textbf{56.7}_{\pm0.21}$ & $\textbf{50.7}_{\pm0.49}$ & $\textbf{43.9}_{\pm0.57}$ \\ \hline
    \end{tabular}
    \label{tab:cifar100all}
\end{table*}

\noindent \textbf{The dilemma with unlabeled data in class-imbalanced SSL with decoupled training.} 
Previous works \cite{kang2019decoupling, zhou2020bbn} suggested that it is better not to apply data re-sampling when learning the representation in a fully supervised learning setup. 
However, in the class-imbalanced semi-supervised learning setting it remains controversial as the role of the unlabeled data needs to be considered. 
On the one hand, plenty of works \cite{sohn2020fixmatch, chen2020big, chen2020improved} have shown that large amounts of unlabeled data can effectively help the model to learn better models when there is only limited labeled data or even when there is no labeled data available. This demonstrates the positive value of the unlabeled data. 
On the other hand, our experiments show that if we adopt a random sampling strategy for the labeled data, it will inevitably lead to biased pseudo labels and harm the quality of the unlabeled data. This will harm the performance of the model. 
The dilemma of unlabeled data can thus be summarised as existing experiences from supervised setting suggest that re-sampling is harmful at the representation learning stage but without re-sampling, the pseudo labels are biased and will further harm the model performance.
Motivated by the intuition that random sampling at the representation learning stage might be sub-optimal, we conduct systematical experiments to analyze the effect of re-sampling at the representation learning stage.

\noindent \textbf{Experimental Setup.} Our experiments build on the jointly trained models with different re-sampling strategies from Table \ref{tab:cifar10joint}.
We follow the idea of decoupling the representation and classifier by freezing the feature extractor and fine-tuning the classifier with a different data re-sampling strategy. Here we use a mean sampler for both labeled data and unlabeled data in the fine-tuning stage to obtain a class-balanced classifier as models fine-tuned with other re-sampling strategies produce similar performance trends and do not violate our conclusion.

\noindent \textbf{Main Results.} Table \ref{tab:cifar10decoupled} summarizes the results on CIFAR-10. We observe that the fine-tuned FixMatch achieves a consistent improvement compared to the jointly trained counterparts in most scenarios. 
Recall that the performance of models without re-sampling (Random/Random) is much lower compared to those with re-sampling in Table \ref{tab:cifar10joint}, and the produced pseudo labels are biased. 
However, after fine-tuning the classifier, these models achieve the most significant performance improvements (see Table \ref{tab:cifar10decoupled} underlined numbers) and obtain the best performance in almost all settings. 
This phenomenon clearly highlights that the biased pseudo labels in the unlabeled data do no harm to the learning of representation. Even more, they improve the quality of the feature extractor, which can be revealed by fine-tuning the classifier with a proper data re-sampling strategy.

\emph{Summary.} When comparing the results of jointly and decoupled trained models (Tables \ref{tab:cifar10joint} \& \ref{tab:cifar10decoupled}), we observe that \textbf{re-sampling only helps to learn the classifier but harms the representation}. Therefore, different data sampling strategies are required in those two learning stages.

\begin{table*}[ht]
    \centering
    \small
    \setlength{\abovecaptionskip}{0pt}   
    \setlength{\belowcaptionskip}{10pt}
    \caption{Per-class classification recall (\%) on CIFAR-10 with imbalanced ratio $\lambda=2$ and unlabeled ratio $\beta=100$. Relative performance comparison is shown in gray.}
    \begin{tabular}{|c|cccccccccc|c|}
        \hline
         & \multicolumn{11}{c|}{Class Index} \\ \cline{2-12} 
        \multirow{-2}{*}{SSL Algorithm} & 1 & 2 & 3 & 4 & 5 & 6 & 7 & 8 & 9 & 10 & Avg \\ \hline
        FixMatch & 98.3 & 99.3 & 87.8 & 80.4 & 88.5 & 65.7 & 75.9 & 58.9 & 46.8 & 37.2 & 73.9 \\
        FixMatch + BiS & 94.4 & 98.0 & 86.1 & 78.7 & 89.4 & 74.2 & 81.3 & 69.4 & 69.9 & 68.6 & 81.0 \\
        \rowcolor[HTML]{C0C0C0} 
         & -3.9 & -1.3 & -1.7 & -1.7 & +0.9 & +8.5 & +5.4 & +10.5 & +23.1 & +31.4 & +7.1 \\ \hline
    \end{tabular}
    \label{tab:per-c}
\end{table*}

\section{Bi-Sampling}
\label{sec:bis}
As suggested before, the biased pseudo labels does not harm the model ability for learning good representations. 
Based on this interesting finding, we suggest to re-think the current paradigm of having a single data sampling strategy when training models for class-imbalanced semi-supervised learning. 
To exploit this finding, we propose a simple yet highly effective Bi-Sampling (BiS) method which adopts two data samplers with different sampling strategies to train a good feature extractor and classifier. BiS builds on the advantages of decoupled training and takes a step further to integrate this decoupling process into an end-to-end framework. 

Formally, at each training step, we use two samplers $A, B$, which will give a different probability $\mu_{Aj}, \mu_{Bj}$ for sampling a data point with label $j$ as indicated before. 
To combine the result of the two samplers, we introduce a ratio parameter $\alpha \in [0,1]$ that controls the weights of two samplers. Thus the probability of sampling a data point with label $j$ to the current training batch is then: 
\begin{align*}
    \mu_{j} = \alpha * \mu_{Aj} + (1 - \alpha) * \mu_{Bj}.
\end{align*}
And that of adding an unlabeled sample with pseudo label $j$ becomes ${\mu_j}^q$.
The ratio $\alpha$ progressively changes during the training process from $1$ to $0$. Concretely, when denoting the total training epochs as $T_{max}$ and the current training epoch as $T$, $\alpha$ is calculated by:
\begin{align*}
    \alpha = 1 - (\frac{T}{T_{max}})^2.
\end{align*}
Intuitively, with gradually decreasing $\alpha$, the focus of the network gradually changes from sampling data that benefits learning a discriminative feature representation into sampling data that helps to learn a well-balanced classifier to improve performance on long-tailed recognition. And since only at the last part of the training process the model tends to rely heavily on the re-sampling sampler, it has a very limited bad effect on the feature extractor thus achieves state-of-the-art performances.
Different from decoupled fine-tuning strategies \cite{kang2019decoupling}, the usage of $\alpha$ ensures the representation and classifier can be updated consistently instead of ignoring the other while training for one goal. Besides, unlike two-branch network design strategies \cite{zhou2020bbn}, our BiS avoids introducing any additional network parameters and requires no further complicated mixup training strategy.

\begin{figure*}[ht]
    \centering
    \includegraphics[width=\linewidth,height=5.5cm]{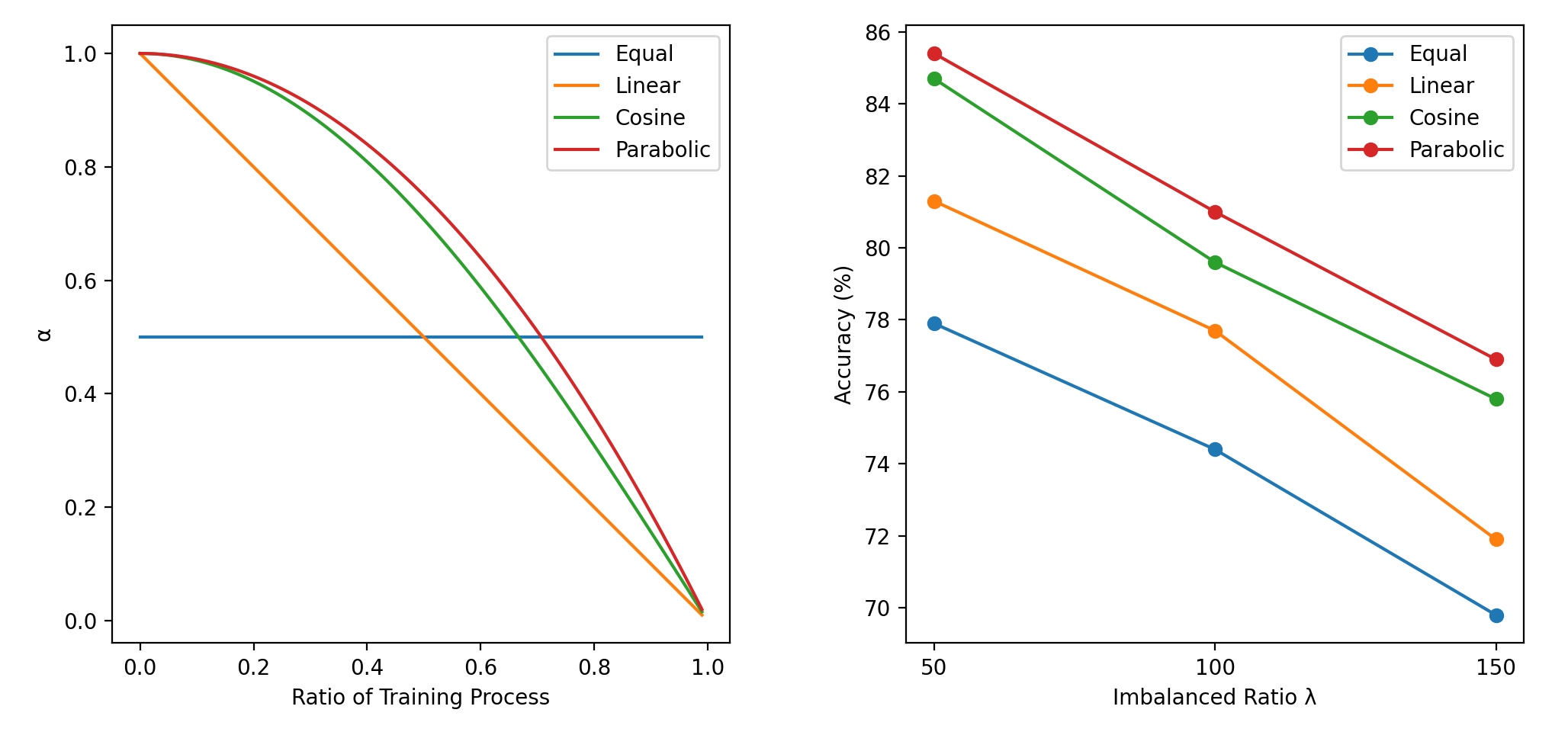}
    \caption{Ablation study on decay strategies. \textbf{Left}: Four different decay schedules for $\alpha$ as a function of the training process. Note how different strategies reach the value of $\alpha=0.5$ at different times in the training process. \textbf{Right}: Performance comparison on class-imbalanced semi-supervised CIFAR-10 with unlabeled ratio $\beta = 2$.}
    \label{fig:func}
\end{figure*}

\begin{figure*}[ht]
    \centering
    \includegraphics[width=\linewidth,height=5.5cm]{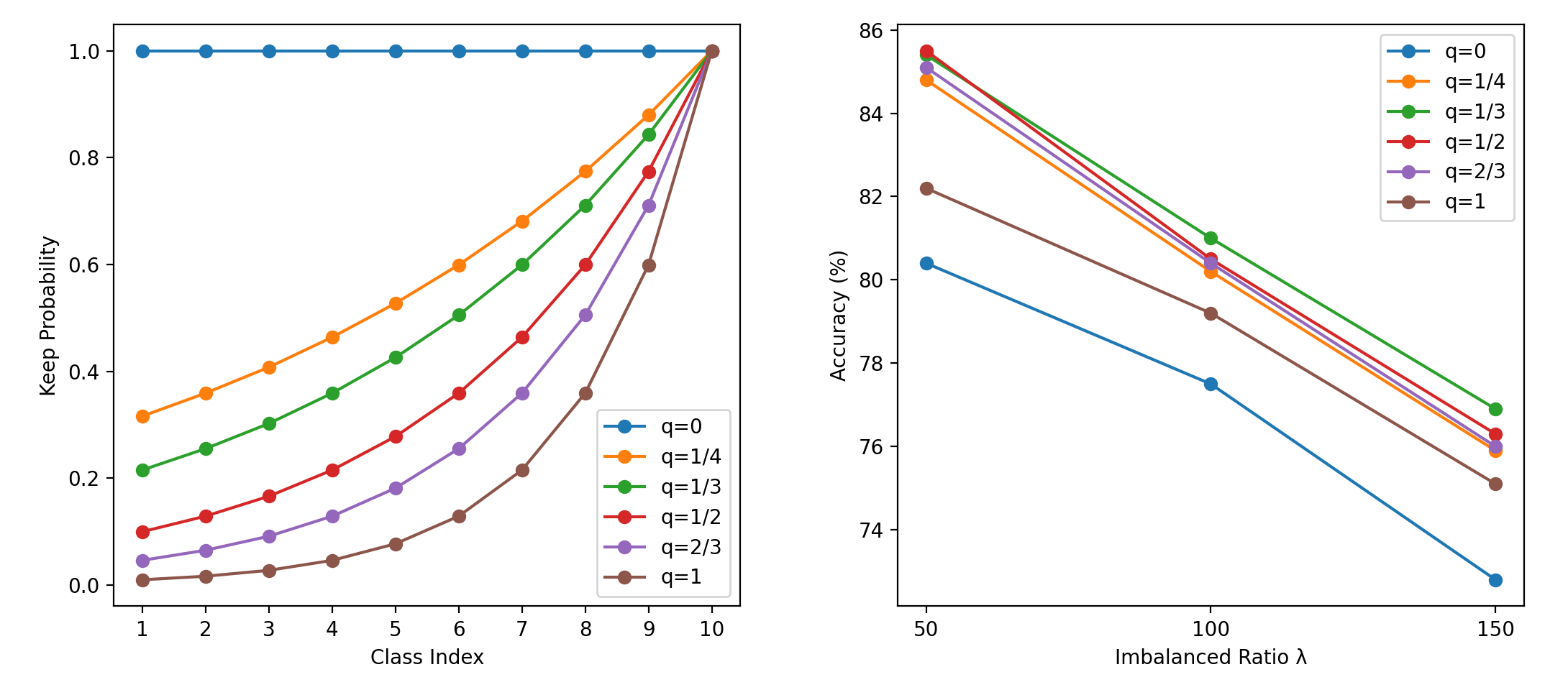}
    \caption{Ablation study on the keep probability $q$. \textbf{Left}: Keep probability of each class with mean sampler under imbalanced ratio $\lambda=100$ of each value. \textbf{Right}: Performance comparison on class-imbalanced semi-supervised CIFAR-10 with unlabeled ratio $\beta = 2$.}
    \label{fig:esp}
\end{figure*}

\subsection{Experiments: Class-imbalanced Semi-Supervised Learning with Bi-Sampling}

\textbf{Experimental Setup.} We choose random sampler for both labeled and unlabeled data as the first sampling strategy and mean sampler for both as the second in our experiments. We compare BiS with recent methods CReST+ \cite{wei2021crest} and DARP \cite{kim2020distribution}. For fair comparison, we follow the paper to combine DARP with the classifier re-training (cRT) algorithm to adapt to long-tailed setting. Both MixMatch \cite{berthelot2019mixmatch} and FixMatch \cite{sohn2020fixmatch} are chosen as baseline algorithms. Average results over 5 different folds are reported to avoid variance.

\noindent \textbf{CIFAR-10.} We report the performance of different methods on CIFAR-10 in Table \ref{tab:cifar10all}. Our BiS achieves state-of-the-art performance in all settings. In particular, for MixMatch our BiS sampling strategy achieves roughly $6\%$ improvement compared to the baseline and $1\%$ improvement compared to CReST+. For FixMatch, the absolute performance gain compared to the baseline and CReST+ is further increased to $7.5\%$ and $2\%$ respectively.

\noindent \textbf{CIFAR-100.} Table \ref{tab:cifar100all} summarizes the performance on class-imbalanced semi-supervised CIFAR-100. Again, our proposed BiS strategy achieves top performance under various settings, providing around $4\%$ and $1\%$ absolute gain compared to FixMatch and CReST+ respectively.

\noindent \textbf{Per-class recall.} Table \ref{tab:per-c} summarizes the per-class recall on CIFAR-10 of FixMatch w/ \& w/o BiS when $\lambda=2$ and $\beta=100$. Our Bi-Sampling strategy achieves significant gains on the two most minority classes ($31.4\%$ and $23.1\%$) while having only little loss the most majority classes, which causes the overall performance improvement of $7.1\%$.

\noindent \textbf{Ablation Study on Decay Strategies.} To better understand the effect of our progressive change between two different data samplers, we conduct an ablation study on the potential decay strategies using the class-imbalanced semi-supervised CIFAR-10 dataset. We use the setting of unlabeled ratio $\beta = 2$ with varying imbalanced ratio $\lambda$. We adopt four methods to calculate $\alpha$ as follows: Equal ($\alpha=0.5$), Linear ($\alpha=1-\frac{T}{T_{max}}$), Cosine ($\alpha=\cos{(\frac{T}{T_{max}} \cdot \frac{\pi}{2}})$), Parabolic ($\alpha=1-(\frac{T}{T_{max}})^2$). The shape of each function performance comparisons is shown in Figure \ref{fig:func}. We observe that the slower the function decays to $0$, the better the performance is. An equal sampling gives the worst performance, while our adopted parabolic decay improves around $8\%$ compared to the equal weighting.

\noindent \textbf{Ablation Study on Keep Probability $\boldsymbol{q}$.} To better evaluate the effect of different re-sampling strength on the final performance, we conduct an ablation study on the keep probability $q$ of unlabeled data on the CIFAR-10 dataset under the setting of unlabeled ratio $\beta = 2$ with varying imbalanced ratio $\lambda$. 
We choose six values for $q$. As $q$ decreases, the re-sampling strength also gets weaker. Figure \ref{fig:esp} shows the performance comparison of different values for $q$. If the value of $q$ is too low (i.e. $q=0$) or too high (i.e. $q=1$) the performance significantly decreases, while the remaining values provide similar results.
\section{Conclusion}

In this work, we studied the effect of data re-sampling strategies in class-imbalanced semi-supervised learning and made several interesting observations. 
We found that although long-tailed data distributions will lead to biased pseudo labels of the unlabeled data during SSL, these wrongly annotated samples do not harm the model when learning the feature extractor. 
On the contrary, accounting for the class-imbalanced through data re-sampling should be avoided as it harms the learning of the feature extractor. 
But on the other hand, data re-sampling was critical for training a well-balanced classifier. 
Based on these findings, we suggested to re-think the current paradigm of having only a single data re-sampling strategy in class-imbalanced SSL, and developed a simple yet highly effective Bi-Sampling training strategy which achieved state-of-the-art performance.

{\small
\bibliographystyle{ieee_fullname}
\bibliography{egbib.bib}

\begin{thebibliography}{10}\itemsep=-1pt

\bibitem{abuelhaija2016youtube8m}
Sami Abu-El-Haija, Nisarg Kothari, Joonseok Lee, Paul Natsev, George Toderici,
  Balakrishnan Varadarajan, and Sudheendra Vijayanarasimhan.
\newblock Youtube-8m: A large-scale video classification benchmark, 2016.

\bibitem{bachman2014learning}
Philip Bachman, Ouais Alsharif, and Doina Precup.
\newblock Learning with pseudo-ensembles.
\newblock {\em arXiv preprint arXiv:1412.4864}, 2014.

\bibitem{berthelot2019remixmatch}
David Berthelot, Nicholas Carlini, Ekin~D Cubuk, Alex Kurakin, Kihyuk Sohn, Han
  Zhang, and Colin Raffel.
\newblock Remixmatch: Semi-supervised learning with distribution alignment and
  augmentation anchoring.
\newblock {\em arXiv preprint arXiv:1911.09785}, 2019.

\bibitem{berthelot2019mixmatch}
David Berthelot, Nicholas Carlini, Ian Goodfellow, Nicolas Papernot, Avital
  Oliver, and Colin Raffel.
\newblock Mixmatch: A holistic approach to semi-supervised learning.
\newblock {\em arXiv preprint arXiv:1905.02249}, 2019.

\bibitem{cao2019learning}
Kaidi Cao, Colin Wei, Adrien Gaidon, Nikos Arechiga, and Tengyu Ma.
\newblock Learning imbalanced datasets with label-distribution-aware margin
  loss.
\newblock In {\em Advances in Neural Information Processing Systems}, 2019.

\bibitem{chen2020big}
Ting Chen, Simon Kornblith, Kevin Swersky, Mohammad Norouzi, and Geoffrey
  Hinton.
\newblock Big self-supervised models are strong semi-supervised learners, 2020.

\bibitem{chen2020improved}
Xinlei Chen, Haoqi Fan, Ross Girshick, and Kaiming He.
\newblock Improved baselines with momentum contrastive learning, 2020.

\bibitem{cui2019class}
Yin Cui, Menglin Jia, Tsung-Yi Lin, Yang Song, and Serge Belongie.
\newblock Class-balanced loss based on effective number of samples.
\newblock In {\em Proceedings of the IEEE/CVF Conference on Computer Vision and
  Pattern Recognition}, pages 9268--9277, 2019.

\bibitem{grandvalet2005semi}
Yves Grandvalet, Yoshua Bengio, et~al.
\newblock Semi-supervised learning by entropy minimization.
\newblock In {\em CAP}, pages 281--296, 2005.

\bibitem{he2015deep}
Kaiming He, Xiangyu Zhang, Shaoqing Ren, and Jian Sun.
\newblock Deep residual learning for image recognition, 2015.

\bibitem{hyun2020class}
Minsung Hyun, Jisoo Jeong, and Nojun Kwak.
\newblock Class-imbalanced semi-supervised learning.
\newblock {\em arXiv preprint arXiv:2002.06815}, 2020.

\bibitem{johnson2008accuracy}
C~Daniel Johnson, Mei-Hsiu Chen, Alicia~Y Toledano, Jay~P Heiken, Abraham
  Dachman, Mark~D Kuo, Christine~O Menias, Betina Siewert, Jugesh~I Cheema,
  Richard~G Obregon, et~al.
\newblock Accuracy of ct colonography for detection of large adenomas and
  cancers.
\newblock {\em New England Journal of Medicine}, 359(12):1207--1217, 2008.

\bibitem{kang2019decoupling}
Bingyi Kang, Saining Xie, Marcus Rohrbach, Zhicheng Yan, Albert Gordo, Jiashi
  Feng, and Yannis Kalantidis.
\newblock Decoupling representation and classifier for long-tailed recognition.
\newblock {\em arXiv preprint arXiv:1910.09217}, 2019.

\bibitem{kim2020distribution}
Jaehyung Kim, Youngbum Hur, Sejun Park, Eunho Yang, Sung~Ju Hwang, and Jinwoo
  Shin.
\newblock Distribution aligning refinery of pseudo-label for imbalanced
  semi-supervised learning.
\newblock {\em arXiv preprint arXiv:2007.08844}, 2020.

\bibitem{kim2020m2m}
Jaehyung Kim, Jongheon Jeong, and Jinwoo Shin.
\newblock M2m: Imbalanced classification via major-to-minor translation.
\newblock In {\em Proceedings of the IEEE/CVF Conference on Computer Vision and
  Pattern Recognition}, pages 13896--13905, 2020.

\bibitem{krizhevsky2009learning}
Alex Krizhevsky, Geoffrey Hinton, et~al.
\newblock Learning multiple layers of features from tiny images.
\newblock 2009.

\bibitem{lee2013pseudo}
Dong-Hyun Lee et~al.
\newblock Pseudo-label: The simple and efficient semi-supervised learning
  method for deep neural networks.
\newblock In {\em Workshop on challenges in representation learning, ICML},
  volume~3, 2013.

\bibitem{lin2015microsoft}
Tsung-Yi Lin, Michael Maire, Serge Belongie, Lubomir Bourdev, Ross Girshick,
  James Hays, Pietro Perona, Deva Ramanan, C.~Lawrence Zitnick, and Piotr
  Dollár.
\newblock Microsoft coco: Common objects in context, 2015.

\bibitem{Liu_2020_CVPR}
Jialun Liu, Yifan Sun, Chuchu Han, Zhaopeng Dou, and Wenhui Li.
\newblock Deep representation learning on long-tailed data: A learnable
  embedding augmentation perspective.
\newblock In {\em Proceedings of the IEEE/CVF Conference on Computer Vision and
  Pattern Recognition (CVPR)}, June 2020.

\bibitem{long2015fully}
Jonathan Long, Evan Shelhamer, and Trevor Darrell.
\newblock Fully convolutional networks for semantic segmentation, 2015.

\bibitem{mahajan2018exploring}
Dhruv Mahajan, Ross Girshick, Vignesh Ramanathan, Kaiming He, Manohar Paluri,
  Yixuan Li, Ashwin Bharambe, and Laurens Van Der~Maaten.
\newblock Exploring the limits of weakly supervised pretraining.
\newblock In {\em Proceedings of the European Conference on Computer Vision
  (ECCV)}, pages 181--196, 2018.

\bibitem{mclachlan1975iterative}
Geoffrey~J McLachlan.
\newblock Iterative reclassification procedure for constructing an
  asymptotically optimal rule of allocation in discriminant analysis.
\newblock {\em Journal of the American Statistical Association},
  70(350):365--369, 1975.

\bibitem{ren2016faster}
Shaoqing Ren, Kaiming He, Ross Girshick, and Jian Sun.
\newblock Faster r-cnn: Towards real-time object detection with region proposal
  networks, 2016.

\bibitem{russakovsky2015imagenet}
Olga Russakovsky, Jia Deng, Hao Su, Jonathan Krause, Sanjeev Satheesh, Sean Ma,
  Zhiheng Huang, Andrej Karpathy, Aditya Khosla, Michael Bernstein,
  Alexander~C. Berg, and Li Fei-Fei.
\newblock Imagenet large scale visual recognition challenge, 2015.

\bibitem{sajjadi2016mutual}
Mehdi Sajjadi, Mehran Javanmardi, and Tolga Tasdizen.
\newblock Mutual exclusivity loss for semi-supervised deep learning.
\newblock In {\em 2016 IEEE International Conference on Image Processing
  (ICIP)}, pages 1908--1912. IEEE, 2016.

\bibitem{sajjadi2016regularization}
Mehdi Sajjadi, Mehran Javanmardi, and Tolga Tasdizen.
\newblock Regularization with stochastic transformations and perturbations for
  deep semi-supervised learning.
\newblock {\em arXiv preprint arXiv:1606.04586}, 2016.

\bibitem{shen2016relay}
Li Shen, Zhouchen Lin, and Qingming Huang.
\newblock Relay backpropagation for effective learning of deep convolutional
  neural networks.
\newblock In {\em European conference on computer vision}, pages 467--482.
  Springer, 2016.

\bibitem{sohn2020fixmatch}
Kihyuk Sohn, David Berthelot, Chun-Liang Li, Zizhao Zhang, Nicholas Carlini,
  Ekin~D Cubuk, Alex Kurakin, Han Zhang, and Colin Raffel.
\newblock Fixmatch: Simplifying semi-supervised learning with consistency and
  confidence.
\newblock {\em arXiv preprint arXiv:2001.07685}, 2020.

\bibitem{NIPS2017_147ebe63}
Yu-Xiong Wang, Deva Ramanan, and Martial Hebert.
\newblock Learning to model the tail.
\newblock In I. Guyon, U.~V. Luxburg, S. Bengio, H. Wallach, R. Fergus, S.
  Vishwanathan, and R. Garnett, editors, {\em Advances in Neural Information
  Processing Systems}, volume~30. Curran Associates, Inc., 2017.

\bibitem{wei2021crest}
Chen Wei, Kihyuk Sohn, Clayton Mellina, Alan Yuille, and Fan Yang.
\newblock Crest: A class-rebalancing self-training framework for imbalanced
  semi-supervised learning.
\newblock {\em arXiv preprint arXiv:2102.09559}, 2021.

\bibitem{yang2020rethinking}
Yuzhe Yang and Zhi Xu.
\newblock Rethinking the value of labels for improving class-imbalanced
  learning.
\newblock {\em arXiv preprint arXiv:2006.07529}, 2020.

\bibitem{zagoruyko2016wide}
Sergey Zagoruyko and Nikos Komodakis.
\newblock Wide residual networks.
\newblock {\em arXiv preprint arXiv:1605.07146}, 2016.

\bibitem{zbontar2018fastMRI}
Jure Zbontar, Florian Knoll, Anuroop Sriram, Tullie Murrell, Zhengnan Huang,
  Matthew~J. Muckley, Aaron Defazio, Ruben Stern, Patricia Johnson, Mary Bruno,
  Marc Parente, Krzysztof~J. Geras, Joe Katsnelson, Hersh Chandarana, Zizhao
  Zhang, Michal Drozdzal, Adriana Romero, Michael Rabbat, Pascal Vincent,
  Nafissa Yakubova, James Pinkerton, Duo Wang, Erich Owens, C.~Lawrence
  Zitnick, Michael~P. Recht, Daniel~K. Sodickson, and Yvonne~W. Lui.
\newblock {fastMRI}: An open dataset and benchmarks for accelerated {MRI}.
\newblock 2018.

\bibitem{zhong2019unequal}
Yaoyao Zhong, Weihong Deng, Mei Wang, Jiani Hu, Jianteng Peng, Xunqiang Tao,
  and Yaohai Huang.
\newblock Unequal-training for deep face recognition with long-tailed noisy
  data.
\newblock In {\em Proceedings of the IEEE/CVF Conference on Computer Vision and
  Pattern Recognition}, pages 7812--7821, 2019.

\bibitem{zhou2020bbn}
Boyan Zhou, Quan Cui, Xiu-Shen Wei, and Zhao-Min Chen.
\newblock Bbn: Bilateral-branch network with cumulative learning for
  long-tailed visual recognition.
\newblock In {\em Proceedings of the IEEE/CVF Conference on Computer Vision and
  Pattern Recognition}, pages 9719--9728, 2020.

\end{thebibliography}
}

\end{document}